\pdfoutput=1
\documentclass[lettersize,journal]{IEEEtran}
\usepackage{amsmath,amsfonts}
\usepackage{array}
\usepackage[caption=false,font=normalsize,labelfont=sf,textfont=sf]{subfig}
\usepackage{textcomp}
\usepackage{stfloats}
\usepackage{url}
\usepackage{verbatim}
\usepackage{graphicx}
\usepackage{cite}
\usepackage{color}
\usepackage{bm}
\usepackage{graphicx}
\usepackage{multirow}
\usepackage{booktabs}
\usepackage{latexsym}
\usepackage{tabularx}
\usepackage{stfloats}
\usepackage{epic}
\usepackage{dashrule}
\usepackage{cuted}\stripsep -3pt
\usepackage{amsmath,lipsum}
\usepackage{makecell}
\usepackage[ruled,linesnumbered]{algorithm2e}
\usepackage{graphicx}
\usepackage{amsmath}
\usepackage{amssymb}
\usepackage{booktabs}
\usepackage{multirow}
\usepackage{caption}
\usepackage{stfloats}
\usepackage{amsmath}
\newcommand{\JS}{\textcolor{red}}

\usepackage{booktabs}
\usepackage{multirow}

\usepackage{stfloats}

\usepackage[ruled,linesnumbered]{algorithm2e}

\begin{document}
\title{SLLEN: Semantic-aware Low-light Image Enhancement Network}
\author{{Mingye Ju, Chuheng Chen, Charles A. Guo, Jinshan Pan,~\IEEEmembership{Member,~IEEE}, \\Jinhui Tang,~\IEEEmembership{Senior Member,~IEEE,} and Dacheng Tao,~\IEEEmembership{Fellow,~IEEE}}
	% <-this % stops a space

		\thanks{This work was supported by Jiangsu Provincial Natural Science Foundation for Higher Education Institutions (23KJB520027), and Research
		Foundation of Nanjing University of Posts and Telecommunications (NY222160).}
		
		\thanks{Mingye Ju is with Nanjing University of Posts and Telecommunications, Nanjing 210000, China, and also with Nanjing University of Science and Technology, Nanjing 210094, China (e-mail: jumingye@njupt.edu.cn).}
		
		\thanks{Chuheng Chen is with Nanjing University of Posts and Telecommunications, Nanjing 210000, China (e-mail: ch.chenchuheng@gmail.com).}
		
		\thanks{Charles A. Guo is with the University of Sydney, 6 Cleveland St, Darlington, NSW 2008, Australia, and also with University of New South Wales, Sydney, NSW 2052, Australia (e-mail: guocharl@gmail.com).}
	
		\thanks{Jinshan Pan and Jinhui Tang are with Nanjing University of Science and Technology, Nanjing 210094, China (e-mail: sdluran@gmail.com; jinhuitang@njust.edu.cn).}
		
		\thanks{Dacheng Tao is with the University of Sydney, 6 Cleveland St, Darlington, NSW 2008, Australia (e-mail: dacheng.tao@sydney.edu.au).}
	
	}% <-this % stops a space<-this % stops a space
%
%\markboth{Sumbitted to IEEE Transactions on Consumer Electronics}
\markboth{Submitted to IEEE Transactions on Computational Imaging
	}{Roberg \MakeLowercase{\textit{et al.}}: High-Efficiency Diode and Transistor Rectifiers}



%\markboth{Journal of \LaTeX\ Class Files,~Vol.~14, No.~8, August~2021}



\maketitle

\begin{abstract}
How to effectively explore semantic feature is vital for Low-light image enhancement (LLE). Existing methods usually utilize the semantic feature that is only drawn from the output produced by high-level semantic segmentation (SS) network. However, if the output is not accurately estimated, it would affect the high-level semantic feature (HSF) extraction, which accordingly interferes with LLE. To this end, we develop a simple and effective semantic-aware LLE network (SLLEN) composed of a LLE main-network (LLEmN) and a SS auxiliary-network (SSaN). In SLLEN, LLEmN integrates the random intermediate embedding feature (IEF), i.e., the information extracted from the intermediate layer of SSaN, together with the HSF into a unified framework for better LLE. SSaN is designed to act as a SS role to provide HSF and IEF. Moreover, thanks to a shared encoder between LLEmN and SSaN, we further propose an alternating training mechanism to facilitate the collaboration between them. Unlike currently available approaches, the proposed SLLEN is able to fully lever the semantic information, e.g., IEF, HSF, and SS dataset, to assist LLE, thereby leading to a more promising enhancement performance. Additionally, the proposed SLLEN can be applied into intelligent transportation system (ITS). The images enhanced by SLLEN are not only visually clear, but also can be better recognized by subsequent semantics. Comparisons between the proposed SLLEN and other state-of-the-art techniques demonstrate the superiority of SLLEN with respect to LLE quality over all the comparable alternatives.
	
\end{abstract}

%%%%%%%%% BODY TEXT
\section{Introduction}
High-quality images are crucial in a large number of computer vision applications, such as tracking \cite{liu2020deep}, segmentation \cite{liu2020location}, detection \cite{yu2020context}, etc. Driven by these needs, a lot of image enhancement methods were proposed in recent years. For example, Zhang \emph{et al.} \cite{10284536} introduced a generative adversarial and self-supervised dehazing network to enhance hazy images. In \cite{9969152}, Huang \emph{et al.} proposed a deep convolutional neural network (CNN)-based blind model for real image denoising. Cui \emph{et al.} \cite{10398593} built a joint local and global representation learning framework for image restoration. Unfortunately, when images are captured under low-light conditions, there is typically impairment by visual distractions or catastrophic loss of information detail. To mitigate this issue, many low-light image enhancement (LLE) methods have been proposed recently.
\par
Currently emerged LLE technologies can be roughly categorized into two groups: model-based methods and learning-based methods.
\par
{\bfseries Model-based Methods:} The most representative model used in LLE methods is Retinex model \cite{retinex1977}. This model assumes that an image $\bf{I}$ can be constructed by the illumination map $\bf{U}$ and reflectance map $\bf{R}$, which can be expressed by:
\begin{equation}
	\bf{I}=\bf{R}\cdot\bf{U},
	\label{eq:Retinex}
\end{equation}
where $\cdot$ denotes element-wise multiplication. According to this model, numerous implementations have been developed \cite{fu2016weighted, guo2016lime, wang2019low, li2018structure, ju2021ide, ng2011total, ren2020lr3m, xu2012structure}. For example, Guo \emph{et al.} \cite{guo2016lime} proposed low-light image enhancement (LIME), which refined the coarse illumination map by imposing a structure prior on Retinex model.
By exploring the relationship between total reflected light and scene depth, Ju \emph{et al.}\cite{ju2021ide} optimized this model and then injected the gray-world assumption on it to expose and dehaze simultaneously.
To eliminate the heavy noise cover in an image, Ren \emph{et al.} \cite{ren2020lr3m} proposed low-rank regularized retinex model, which integrated low-rank prior into a Retinex decomposition process and successfully suppressed remaining noise in the scenes.
In \cite{xu2012structure}, Li \emph{et al.} used a smooth prior to reconstruct LLE problem into an optimization function involving noise distribution. These model-based LLE methods are capable of producing satisfactory results for some cases. However, as the hand-craft priors do not fully exploit the properties of clear images, they may induce unrealistic artifacts in the enhanced results, especially when handling images with inhomogeneous illumination.
\par
{\bfseries Learning-based Methods:} Significant progress has been made in learning-based LLE approaches \cite{ren2019low, yao2024gaca, guo2020zero, jiang2021enlightengan, zhang2021beyond, wang2019underexposed, yang2020fidelity, zheng2021adaptive, zhang2023learning,liu2021retinex, xu2022snr, he2023low, wu2022uretinex,  li2023pixel, luo2023pseudo, cui2022progressive,  kar2024self} due to the development of deep learning.
Typically, Ren \emph{et al.} \cite{ren2019low} designed a deep hybrid network for LLE, which consists of two blocks, i.e., one content block is used for estimating the global contrast of the input image, and the other is an edge block used for modeling structure details. Yao \emph{et al.} \cite{yao2024gaca} proposed a gradient-aware and contrastive-adaptive (GACA) learning framework to implement the low light image enhancement.
Guo \emph{et al.}\cite{guo2020zero} built a lightweight convolutional network to enhance low-light images by estimating an image-specific curve in an unsupervised way.
Following a divide-and-conquer principle, Zhang \emph{et al.} \cite{zhang2021beyond} established a deep neural network for simultaneously kindling the darkness, removing the degradation, and providing users a friendly light adjustment.
Zheng \emph{et al.} \cite{zheng2021adaptive} proposed an adaptive unfolding total variation network to remove noise and compensate lighting for low-light images. Zhang \emph{et al.} \cite{zhang2023learning} proposed a streamlined single convolution layer model (SCLM) to provide global low-light enhancement as the coarsely enhance results.
In \cite{xu2022snr}, an SNR-aware framework was presented, its key contribution is to employ signal-to-noise-ratio-aware transformers and convolutional models to achieve image enhancement. To suppress noise in complex lighting conditions, He \emph{et al.} \cite{he2023low} designed a multiscale illumination adjustment network with a wavelet-based attention mechanism.
Inspired by Retinex model, Wu \emph{et al.} \cite{wu2022uretinex} unfolded the LLE task into a learnable network, which decomposed a low-light image into illumination and scene reflectance to recover high-quality scenes. Kar \emph{et al.} \cite{kar2024self} proposed a novel unpaired low-light image enhancement network that employs novel controlled transformation-based self-supervision and unpaired self-conditioning strategies to address the difficulty of generalising supervised methods to a variety of real-world images.

%Inspired by Retinex model, Wu \emph{et al.} \cite{wu2022uretinex} unfolded an optimization problem into a learnable network, which decomposed a low-light image into illumination and reflectance to better enhance low-light images.
%

These networks are able to realize remarkable performance on most low-light datasets. Nevertheless, they cannot reliably correct the lightness distribution in either bright or dark regions, i.e., unable to work well on images with inhomogeneous illumination.% This is because the different regionsing of inhomogeneous lighting images have different exposure level.
% can hardly be handled by the methods to enhance the low-light image as a whole.

Recently, a few semantic-guided learning-based efforts \cite{liang2022semantically, fan2020integrating, TIPsemantic, 9812457} have been devoted to solving such inhomogeneous pollution issue. To the best of our knowledge, they are all based on the output map of semantic segmentation (SS) network to explore the interaction between low-level feature and semantic feature. For instance, Liang \emph{et al.} \cite{liang2022semantically} imported the enhanced image into SS network to produce a semantic map, and then combined this map and its ground truth to design a semantic brightness loss to assist the training of LLE network. Another solution advocated by \cite{fan2020integrating, TIPsemantic} is based on semantic information extracted from the output map of SS network to guide the adjustment of low-level feature. The main advantage is that it does not require the ground truth of SS map to participate during training, which makes it easier to implement compared to \cite{liang2022semantically}. Although this kind of algorithm can alleviate the limitation of inhomogeneous illumination to some extent, their results are still prone to inevitable poor visual effects. This is since SS network might lose efficacy for some cases, which makes its outputs impossible to provide the accurate semantic feature.

\begin{figure*}[ht]
	\centering
	\includegraphics[scale=0.30]{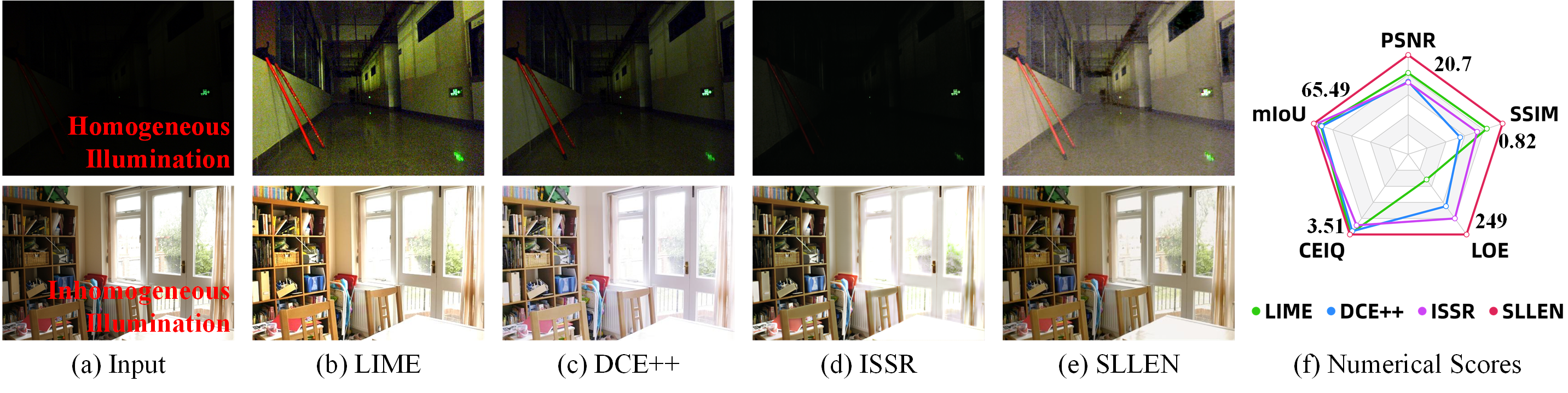}
	\caption{The comparisons between the proposed SLLEN and three most representative state-of-the-art techniques, including LIME (model-based) \cite{guo2016lime}, DCE++ (learning-based) \cite{9369102}, and ISSR (semantic-guided learning-based) \cite{fan2020integrating}. (a)$\sim$(e): Visual comparison on two given examples; As shown, the results of the proposed SLLEN are superior in terms of contrast, exposure, and color, while the results of others are either under-enhanced or over-enhanced. (f): The average scores of different methods in terms of PSNR, SSIM, LOE, CEIQ, and mIoU on several commonly-used datasets remarked in Subsections~\ref{section3-2} and~\ref{section3-3}; it can be easily noticed that our SLLEN remarkably outperforms these competitors.}
	\label{figure1}
\end{figure*}

\begin{figure*}[ht]
	\centering
	\includegraphics[scale=1]{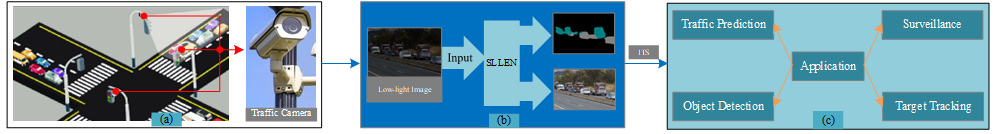}
	\caption{The system architecture for applying the proposed SLLEN to ITS. (a) is an intersection deployed with traffic cameras. (b) is the low-light image enhancement procedure of our SLLEN. (c) is the applications of enhanced images in ITS}
	\label{ITS}
\end{figure*}

{\bfseries Our motivation and contributions:}
In summary, currently available LLE algorithms usually lack the ability to effectively handle images under inhomogeneous illumination conditions, even with the help of high-level semantic features (HSF) fixedly extracted from the output map of SS network. 
% Actually, random semantic information may perform better than such fixed HSF, which has been proved in previous related research, e.g., the performance of random membership degree based fuzzy C-means \cite{dunn1973fuzzy} is clearly superior to that of fixed centers based K-means.
To fully utilize semantic features, a semantic-aware LLE network (SLLEN) composed of an LLE main-network (LLEmN) and an SS auxiliary-network (SSaN) is developed. 
Its core idea is to inject semantic information from SSaN into LLEmN, and then use a shared encoder to do a cooperation between LLEmN and SSaN, thereby ensuring a high-quality semantic-aware LLE. It is worth mentioning that such structure also offers our SLLEN a better potential to be used for traffic object enhancement in ITS, such as object detection, target tracking, surveillance, and traffic prediction, as shown in Figure~\ref{ITS}. The main contributions of this paper are as follows:

\begin{itemize}
\item[$\bullet$]  We propose a fully interactive network (SLLEN) between LLE task (LLEmN) and SS task (SSaN). As the mai-network, LLEmN not only employs the HSF extracted from SS map, but also utilizes random semantic information, i.e., the intermediate embedding feature (IEF) grasped from the intermediate layer in SSaN, to learn more possibilities of light exposure of different regions. As a result, our SLLEN generalizes well to inhomogeneous illumination condition.

%to accurately preserve the information in bright regions and light up the details in dim parts
%under the guidance of both the fixed HSF and the random IEF.

% $\bullet$ We utilize an attention mechanism to introduce HSF into low-level feature and estimate two vectors from IEF to adjust low-level feature using a non-linear manner.
%The benefit of doing so is that it can make up for the deficiency between fixed HSF and random IEF mutually, thereby ensuring more accurate semantic guidance for low-level restoration tasks.

\item[$\bullet$]  Benefiting from a shared encoder connecting the LLEmN and SSaN, we design an alternating training mechanism that iteratively trains the main-network and auxiliary-network by repeatedly tuning the parameters. Such a mechanism further facilitates the interaction between LLEmN and SSaN, and makes that the shared encoder, i.e., encoder of LLEmN, is capable of capturing LLE and SS attributes at the same time, which is very useful for semantic-aware LLE.

% $\bullet$ We devise a knowledge distillation loss, illumination total variation loss, and gradient loss, to constrain the SLLEN for better performance.

% We design an effective image enhancement strategy with five loss function to promote the learning ability of ACB-Net.

\item[$\bullet$]  Extensive experiments demonstrate that the proposed SLLEN achieves favorable performance against state-of-the-art approaches in terms of visual quality and quantitative scores, as a brief description in Figure~\ref{figure1}. Moreover, compared to existing works, the results enhanced by SLLEN are more suitable for information extraction of subsequent high-level vision systems. 
\end{itemize}

\begin{figure*}[ht]
	\centering
	\includegraphics[scale=0.60]{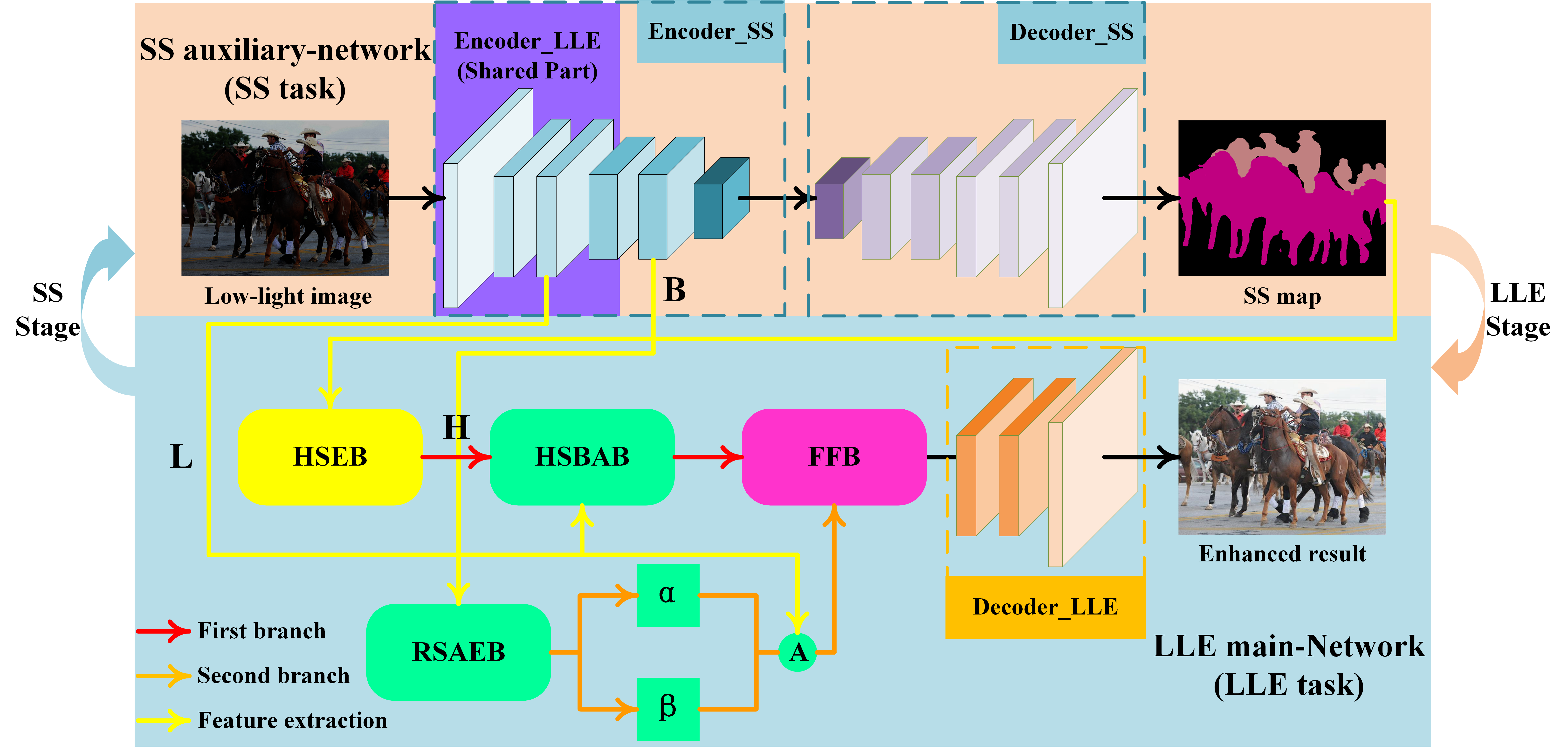}
	\caption{The architecture of the proposed SLLEN. It consists of LLE main-network (LLEmN) and SS auxiliary-network (SSaN), where LLEmN contains Feature Extraction Module (highlight in yellow), Feature Enhancement Module (highlight in green), Feature Fusion Module (highlight in purple).}
		% Two enhancement branches are included in this architecture, which are highlighted in red and blue, respectively.}
	\label{overall_network}
\end{figure*}

\section{Methodology}

% Recall that in the above, intermediate embedding feature (IEF) contains some random characteristics on semantic compared to high-level semantic feature (HSF), which makes it able to more flexibly leverage the hidden interrelation between low-level feature and semantic feature. According to this fact, a semantic-aware image enhancement network (SLLEN) is proposed to learn the mapping between low-light images and clear scenes. The baseline of our network is U-Net ~\cite{2015U}, and only three modules are consisted in this network, i.e., Feature Extraction Module, Feature Enhancement Module, and Decoder Module. The overall architecture of the SLLEN is illustrated in Figure~\ref{overall_network}. As shown, both IEF and HSF are separately embedded into different branches to strengthen our network. Because the Decoder Module is the same as the general encoder-decoder framework ~\cite{2015U}, thus we only give a detailed introduction to the first two modules in the following.

Recall that in the above, current LLE algorithms fail to make full use of semantic features, which leads to the ineffectiveness in handling the inhomogeneously illuminated low-light images. To solve this issue, we propose a semantic-aware LLE network (SLLEN) composed of an LLE main-network (LLEmN) and a semantic segmentation (SS) auxiliary-network (SSaN).
Specifically, by embedding semantic information, the LLEmN is implemented to learn the mapping between low-light images and clear scenes; and SSaN is used to provide both intermediate embedding feature (IEF) and high-level semantic feature (HSF) for LLEmN.
To further facilitate the interaction between the SS task and the LLE task, an alternating training mechanism is designed, which enables SLLEN to accomplish LLE task with the assistance of SS pattern. The architecture of the proposed SLLEN is shown in Figure~\ref{overall_network}. It can be observed from this figure that main-network and auxiliary-network have a shared part in the encoder, i.e., $encoder_{LLE}$, whose function is to bridge the gap between them.  
%The baseline of our network is U-Net ~\cite{2015U}, as shown in the Figure~\ref{overall_network}.
%As shown, the main network (LLEmN) contains an LLE encoder $encoder_{LLE}$ and decoder $decoder_{LLE}$ (see the dashed yellow frame in Figure~\ref{overall_network}), and the sub-network (SSAS) contains an SS encoder $encoder_{SS}$ and decoder $decoder_{SS}$ (see the dashed red frame in Figure~\ref{overall_network}). 
%Here we remark that $encoder_{LLE}$ is the shared part of LLEmN and SSAS, which can bridge the gap between the two sub-networks.
 
\subsection{Low-light Image Enhancement main-Network}
As illustrated in Figure~\ref{overall_network}, the auxiliary-network (SSaN) used for SS only contains U-Net encoder $encoder_{SS}$ and decoder $decoder_{SS}$ (see the upper part in Figure~\ref{overall_network}). But, besides the traditional U-Net encoder $encoder_{LLE}$ and decoder $decoder_{LLE}$, there are also three extra modules, i.e., Feature Extraction Module, Feature Enhancement Module, and Feature Fusion Module that are included in the LLEmN (see the lower part in Figure~\ref{overall_network}). Therefore, in this subsection, we will introduce the details of these modules.

%LLEmN is the core component of SLLEN. In addition to basic encoder and decoder structures, three extra modules are consisted in LLEmN, i.e., Feature Extraction Module, Feature Enhancement Module, and Feature Fusion Module. 

%\JS{\subsection{Alternate Mechanism}
%Before introducing the architecture of SLLEN, we present one of the main contributions of this work, which is an alternate mechanism between two computer vision tasks, i.e., image semantic segmentation task and low-light image enhancement task. Such mechanism is proposed to joint two tasks, which thoroughly explores the semantic feature from the semantic segmentation task to guide the LLE task. 
%Generally, a co-encoder is implemented to map the input into the common embedding for two decoders separately for semantic segmentation and LLE. 
%Two different decoder will alternately be trained with the co-encoder, making SLLEN learn the pattern of implementing both two tasks.
%Different from the previous works, SLLEN integrates two independent decoders for different tasks by a co-encoder, which instinctively enables SLLEN to achieve LLE under the guidance of semantic segmentation pattern.
%}

\subsubsection{Feature Extraction Module}
% The Semantic Segmentation Block (SAS) is the basic component of our SLLEN, which is composed of the $co-encoder$ and semantic segmentation decoder $decoder_{SS}$.
% Benefiting from the architecture of U-Net, we extract various semantic feature from the different layers of SAS.
Three kinds of feature, i.e., low-level feature $\bf{L}$, HSF $\bf{H}$, and IEF $\bf{B}$ are extracted in this module. For $\bf{L}$, it is extracted through the last layer of LLE encoder ($encoder_{LLE}$), i.e.,
\begin{equation}
	\bf{L} = \mathit{f_{encoder_{LLE}}}(\bf{I}),
	\label{eq:encoder}
\end{equation}
where $f_{encoder_{LLE}}$ denotes the operation of $encoder_{LLE}$. For $\bf{H}$, it is extracted from SS map $\bf{S}$, i.e., the output of the last layer of SSaN decoder (see Figure~\ref{overall_network}). 
%
%Figure~\ref{extraction_block}(a) gives an example of SS map  $\bf{S}$ of SSaN, i.e., the output of the last layer of SSaN decoder. 
%
To draw $\bf{H}$ from $\bf{S}$, high-level semantic extraction block (HSEB) is designed, which involves three convolutional layers (each layer has 32, 128, and 512 filters of size 3 $\times$ 3) and max pooling with ReLU activation. 
Formally, the process of HSEB can be illustrated by:
\begin{equation}
	\bf{H} = \mathit{f_{HSEB}}(\bf{S}),
	\label{eq:HSEB}
\end{equation}
where $f_{HSEB}$ stands for the operation of the HSEB.

\begin{figure}[htbp]
	\centering
	\includegraphics[scale=0.2350]{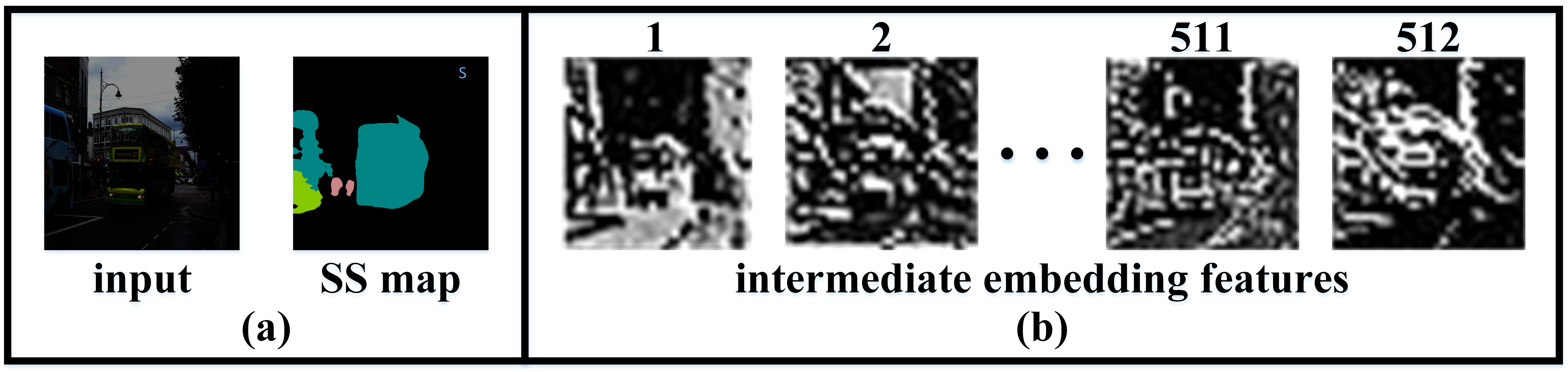}
	\caption{
		%		The details of two blocks in Feature Extraction Module.
		(a): Input low-light image and the corresponding inaccurate SS map produced by SSaN.
		(b): Intermediate embedding features extracted from $5^{th}$ layer of SSaN.
	}
	\label{extraction_block}
\end{figure}

For the last feature $\bf{B}$, here we remark that it is only picked from the intermediate layer of SSaN. This is due to the fact that the feature drawn from the deeper layer in SSaN is closer to $\bf{H}$, while the feature drawn from the shallower layer in SSaN contains more low-level features. In contrast, the intermediate embedding layer could provide some randomness under the condition of maintaining latent semantic feature. Without loss of generality, we pulled out $\bf{B}$ with the size of $512\times16\times16$ from the $5^{th}$ layer of SSaN as the random semantic information. As an example description shown in Figure~\ref{extraction_block}, SSaN fails to produce an accurate semantic map $\bf{S}$ (see Figure~\ref{extraction_block}(a)), which would lead to the errors during extracting high-level semantic feature. Unlike the $\bf{S}$, the grasped $\bf{B}$ has $512$ random semantic maps that are different from each other (see Figure~\ref{extraction_block}(b)), which can feed more randomness to LLEmN and thus promote its generalization ability.

%it can be concluded that the feature $\bf{B}$ containing $512$ random semantic maps has more potential information than SSN's result $\bf{S}$, which can make up for the somehow errors in $\bf{S}$ and is useful to improve \BL{the} generalization ability of our network (The corresponding experiments can be found in Ablation Study (Subsection~\ref{section3-1}) and our supplementary materials).

\begin{figure*}[htbp]
	\centering
	\includegraphics[scale=0.395]{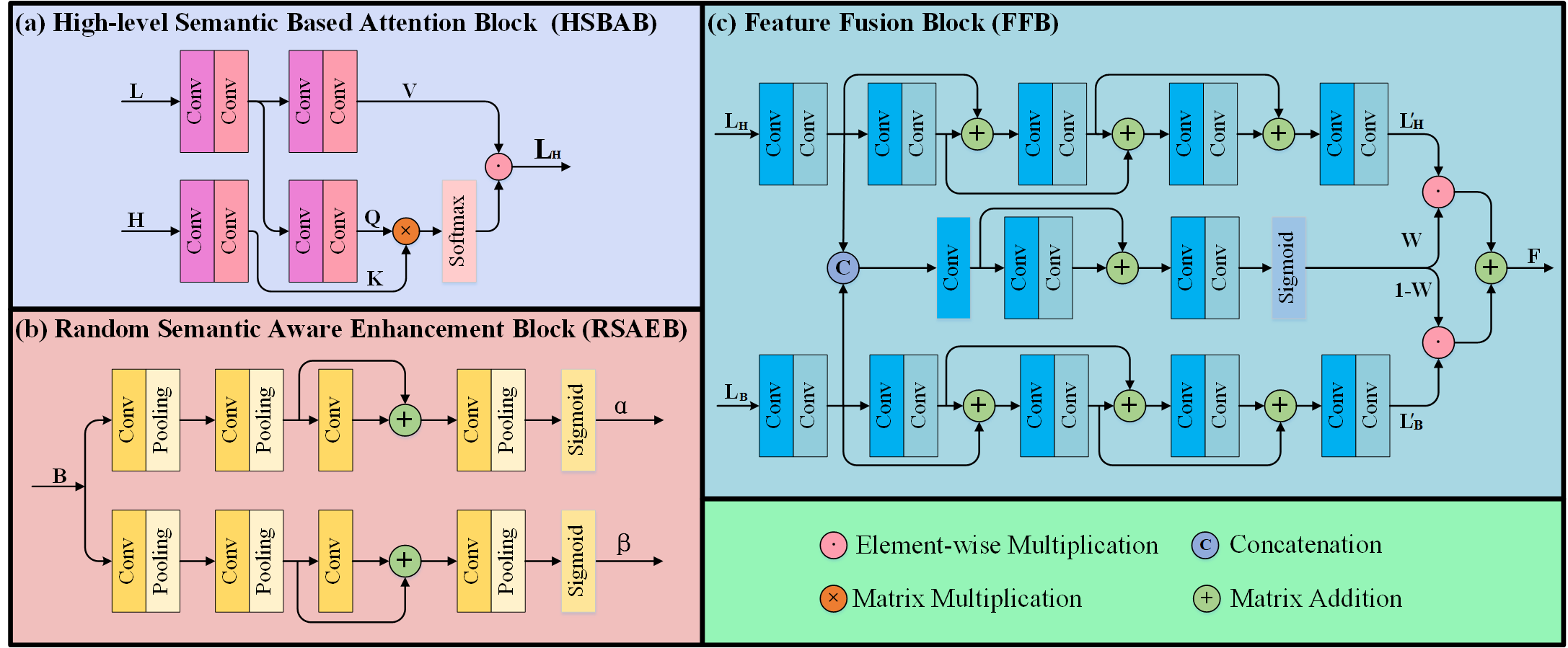}
	\caption{
		(a): The framework of High-level Semantic Based Attention Block (HSBAB).
		(b): The framework of Random Semantic Aware Enhancement Block (RSAEB).
		(c): The framework of Feature Fusion Block (FFB).
	}
	\label{enhancement_block}
\end{figure*}

\subsubsection{Feature Enhancement Module}
Here, we further present a Feature Enhancement Module to enhance low-level feature under the guidance of semantic features.
Figure~\ref{overall_network} describes that this module contains high-level semantic based attention block (HSBAB) and random semantic aware enhancement block (RSAEB), where HSBAB is in the first branch for introducing HSF into low-level feature through an attention mechanism and RSAEB is in the second branch to integrate the random IEF and low-level feature.

\textbf{High-level Semantic Based Attention Block.}
According to the attention mechanism~\cite{NIPS2017_3f5ee243}, vector $\bf{V}$ can be adjusted under the guidance of vector $\bf{K}$ using an attention map generated by mapping its vector $\bf{Q}$ to vector $\bf{K}$.
In this work, the HSBAB is introduced to implement an attention mechanism.
More specifically, as illustrated in Figure~\ref{enhancement_block}(a), $\bf{Q}$ and $\bf{V}$ are generated from $\bf{L}$, and $\bf{K}$ is generated from $\bf{H}$.
In this way, $\bf{L}$ can be effectively adjusted by the generated attention map, which is formulated as:
\begin{equation}
	\bf{L_{H}}=\mathit{SoftMax}(\frac{\bf{Q}\bf{K}^{T}}{\sqrt{d_{k}}}) \bf{V},
	\label{eq:attention}
\end{equation}
where $\bf{L_{H}}$ is the optimized feature of $\bf{L}$ based on HSF $\bf{H}$, $SoftMax(\cdot)$ denotes the SoftMax normalization operation, and $\bf{d_{k}}$ is the normalization factor.
However, because of the drawback of $\bf{H}$ as discussed above, the corresponding result $\bf{L_{H}}$ may fail to sufficiently support the high-quality performance of LLE task (see Subsection~\ref{section3-1}). As a result, random semantic aware enhancement block is proposed as follows.

\textbf{Random Semantic Aware Enhancement Block.}
In \cite{fan2020integrating}, Fan \emph{et al.}~ integrate the HSF into LLE by imposing a linear transformation on low-level feature $\bf{L}$.
Nevertheless, it unintentionally induces two complications; one is that linear transformation is not flexible enough to adjust the $\bf{L}$, and the other is that two semantic based vectors directly calculated from the HSF have the potential to be erroneous due to the fallibility of SS network.
To tackle this problem, instead of using the HSF, we adopt a non-linear transformation on low-level information with the guidance of random IEF $\bf{B}$.
Mathematically, it can be written by:
\begin{equation}
	\bf{L_{B}}=\beta \cdot \bf{L}^{\alpha},
	\label{eq:SPBB}
\end{equation}
where $\bf{L_{B}}$ is the adjusted feature, $\alpha$ and $\beta$ are non-linear transformation vectors with the size of $1 \times 1  \times 512$ generated from IEF $\bf{B}$ using several convolutional layers, as shown in Figure~\ref{enhancement_block}(b).
Compared to the technique \cite{fan2020integrating} that directly employs the HSF extracted from the SS map, our strategy can suppress the limitations of SS network and make use of the random semantic characteristics in IEF to better predict such vectors, which is beneficial to leverage the inherent relationships between low-level feature and semantic feature.

%\BL{As shown in Figure~\ref{extraction_block}(a), intermediate feature maps are different from each other, which provides more possibilities to help the prediction of vectors $\alpha$ and $\beta$.}
%
% Thus the predicted vectors $\alpha$ and $\beta$ are more robust.
%
%This helps to leverage the inherent relationships between low-level feature and semantic feature, thereby achieving better generalization ability for SLLEN.

\subsubsection{Feature Fusion Module}Once $\bf{L_{H}}$ and $\bf{L_{B}}$ have been computed, we adopt a feature fusion block (FFB) to aggregate the high-level semantic aware feature $\bf{L_{H}}$ and semantic aware feature $\bf{L_{B}}$.
Figure~\ref{enhancement_block}(c) depicts the detailed structure of the FFB. It can be concluded from this figure that $\bf{L_{H}}$ and $\bf{L_{B}}$ can be fused through several convolution operators, and weight $\bf{W}$ is used to make a trade-off between $\bf{L_{H}}$ and $\bf{L_{B}}$. Accordingly, this process is expressed as:
\begin{equation}
	\bf{F} = \bf{W} \cdot \bf{L'_{H}}+(\bf{1}-\bf{W}) \cdot \bf{L'_{B}},
	\label{eq:fusion}
\end{equation}
where $\bf{F}$ is the fused feature, $\bf{L'_{H}}$ and $\bf{L'_{B}}$ are the convolutional results in FFB.
Using this strategy, the two semantic-guided low-level features can be well interacted to boost the generalization performance of the proposed LLEmN.
Finally, we reconstruct the enhanced result using fused feature $\bf{F}$ by U-Net decoder $decoder_{LLE}$, as demonstrated in Figure~\ref{overall_network}.

\subsection{Alternating Training Mechanism}
% The training strategy in previous works rarely focused on two training tasks, and even if they do, there is no interaction between them.
% The existing 

In previous sections, we have illustrated the overall architecture of the proposed SLLEN and its working principle. However, how to effectively train this network is also the key to achieve high-quality semantic-aware LLE. In this work, we design an alternating training mechanism, whose core idea is to iteratively train the SSaN and LLEmN in SLLEN, to ensure efficient interaction of semantic segmentation with LLE. For ease of description, two repeatable training stages, i.e., SS stage and LLE stage, are defined in this mechanism. Specifically, for SS stage, we first train SSaN for 1 batch and update its network parameters according to SS samples, enabling the SSaN to roughly extract and provide semantic feature for LLEmN. Then, for LLE stage, we train the LLEmN for the same batch and update the corresponding parameters according to LLE samples. When the LLE stage is completed, return to the SS stage until the convergence condition of the training is met. For clarity, the more specific process of alternating training mechanism is described in Algorithm~\ref{algorithm}. Note that LLEmN's encoder $encoder_{LLE}$ is part of SSaN's encoder $encoder_{SS}$, thus using alternating training mechanism to train SLLEN can make $encoder_{LLE}$ have the ability to capture both low-level features and semantic features at the same time, which is beneficial for semantic-aware LLE. Another advantage of using alternating training mechanism is that the SS dateset and LLE dateset employed to train SLLEN can be asymmetric, indicating that the proposed SSLEN can learn more possibilities from two different types of datasets to better deal with LLE task.

\begin{algorithm}
	\caption{Alternating Training Mechanism}\label{algorithm}
	% \KwIn{Guided image $\bm{I}$;}
	%	\textbf{Function: $check()$} return whether all the pixels in the image are traversed;\\
	%	\textbf{function: $extend()$} return whether all the pixels in the image are scanned.\\
	
	\For{round=1; round!=100; round++}
	{
		\For{batch=1; batch++}
		{\textbf{For SS stage}:\\
			Train SSaN and update the corresponding parameters according to SS samples;\\
			\textbf{For LLE stage}:\\
			%\For{epoch=1; epoch!=XX; epoch++}
			Train LLEmN and update the corresponding parameters according to LLE samples \\ (Note that the parameters in shared part $encoder_{LLE}$ will be updated while the rest parameters in SSaN are frozen).}
	}
	%	\KwResult{Partition pixel-sets $S$;}
\end{algorithm}

\subsection{Loss Function}
During training, the loss function used for training SSaN, i.e., the SS stage, is the same as the Ref.~\cite{2015U}.
For LLE stage, the smooth loss $\mathcal L_{s}$ \cite{wald1950statistical} and the perceptual loss $\mathcal L_{vgg}$ \cite{simonyan2014very} are employed to train LLEmN. Moreover, we further exploit a knowledge distillation loss $\mathcal L_{kd}$, an illumination total variation loss $\mathcal L_{itv}$, and a gradient loss $\mathcal L_{g}$, to improve the performance of LLEmN.

\textbf{Knowledge Distillation Loss.} According to~\cite{hinton2015distilling}, the encoder can serve as a teacher network to supervise the decoder (student network), which makes to yield a knowledge distillation loss:
%
%Therefore, we adopt knowledge distillation loss $\mathcal L_{kd}$, which is defined as:
\begin{equation}
	\mathcal L_{kd} = \sum_{i=1}^n MSE(E_i-D_{n-i+1}),
	\label{eq:KD_loss}
\end{equation}
where $n$ is the number of layers in the encoder-decoder framework, $E_i$ and $D_{n-i+1}$ indicate the embedding from the $i$-th encoder layer and the $n-i+1$-th decoder layer, respectively. By minimizing the deep features of each layer, the student network (decoder) can effectively extract the latent knowledge of the teacher network (encoder), thereby improving its decoding performance.
\par
\textbf{Illumination Total Variation Loss.} Based on Retinex theory~\cite{retinex1977}, low-light image can be decomposed into reflectance map $\bf{R}$ and illumination map $\bf{U}$. Here, we take enhanced image $\bf{O}$ as reflectance map $\bf{R}$, thus $\bf{U}$ can be calculated by rewriting Eq.~(\ref{eq:Retinex}) as:
\begin{equation}
	\bf{U} = \frac{\bf{I}}{\bf{O}}.
	\label{eq:ReRetinex}
\end{equation}
In general, illumination map $\bf{U}$ tends to be smooth spatially \cite{retinex1977}. This inspires us to propose an illumination total variation loss $\mathcal L_{itv}$ to quantify the dissimilarity among the adjacent pixels in the illumination map, i.e.,
\begin{equation}
	\mathcal L_{itv}= \dfrac{1}{CHW}\sum_{c=1}^{C}\sum_{h=1}^{H}\sum_{w=1}^{W}\vert(\nabla_{x}\bf{U} _\mathit{c,h,w})^\mathrm{2}+(\nabla_\mathit{y}\bf{U} _\mathit{c,h,w})^\mathrm{2}\vert,
	\label{eq:ill_loss}
\end{equation}
where $C$, $H$ and $W$ represent separately channel, height and width of $\bf{U}$, $\nabla_{x}$ and $\nabla_{y}$ are the horizontal and vertical gradient operations, respectively.
\par
\textbf{Gradient Loss.} Following exposure control loss that proposed in~\cite{guo2020zero}, we design a gradient loss $\mathcal L_{gra}$ to enhance the contours and textures of restored results.
% which is set within $[0.4,0.6]$ in our experiments.
In detail, the loss $\mathcal L_{gra}$ is defined as:
\begin{equation}
	\mathcal L_{gra} = \dfrac{1}{3 \cdot N}\sum_{i=1}^{N} \Vert \bf{O}^\mathit{g}_\mathit{i}-\mathit{G}\Vert_\mathrm{1},
	\label{eq:gra_loss}
\end{equation}
where $\Vert \cdot \Vert_1$ indicates the $\mathcal L_{1}$ norm, $N$ represents the batch size, $\bf{O}^\mathit{g}_\mathit{i}$ is the average gradient value of $i^{th}$ image of current batch. From this equation, it can be concluded that the goal of the gradient loss is to ensure the average gradient value of SLLEN's result to be a desired level $G$.
%
%In this work, we set $G=0.051$.
\par

\par
\textbf{Total Loss.} The total loss function of our network can be expressed as follows:
\begin{equation}
	\mathcal L_{total}=\mathcal L_{s}+\mathcal L_{vgg}+\lambda_{kd} \cdot \mathcal L_{kd}+\lambda_{itv} \cdot \mathcal L_{itv}+\lambda_{gra} \cdot \mathcal L_{gra},
	\label{eq:total_loss}
\end{equation}
where $\lambda_{kd}$, $\lambda_{itv}$, and $\lambda_{gra}$ are the weight parameters, which are empirically set to be $1$, $5$, and $1$.

\section{Performance Evaluation}
In this section, the performance of the proposed SLLEN is evaluated from different perspectives. First, a parameter study was conducted to determine the value of $G$, which is the key to gradient loss $\mathcal L_{gra}$ being effective. Then, the illumination maps calculated from SLLEN's results were evaluated and compared to those obtained by different state-of-the-art techniques. Subsequently, we conduct some ablation studies to analyze the function of each loss, each module, and alternating training mechanism in SLLEN. Finally, various challenging low-light images were selected to quantitatively and qualitatively compare the performance of different algorithms.% Moreover, we also check the ability of the results enhanced by different algorithms on semantic segmentation.

\subsection{Training datasets and Implementation Details}
In our experiments, the public datasets (LOL-train \cite{Chen2018Retinex} and LSRW~\cite{LSRW}) consisting of low-light images with different exposure levels and the semantic segmentation datesets \cite{Everingham10} were employed to train SSaN and LLEmN, respectively. 
We implement our framework with PyTorch on an NVIDIA 3090Ti GPU.
During training, the batch size and the learning rate applied are initialized as $6$ and $1\times 10^{-4}$, respectively.
The convolution operators used in SLLEN are all set to be $3 \times 3$.
ADAM optimizer is utilized with default parameters for the training of SLLEN.

\subsection{Statistical Results on Expected Gradient Level $G$}
\label{section1}
\begin{figure}[h]
	\centering
	\includegraphics[scale=0.65]{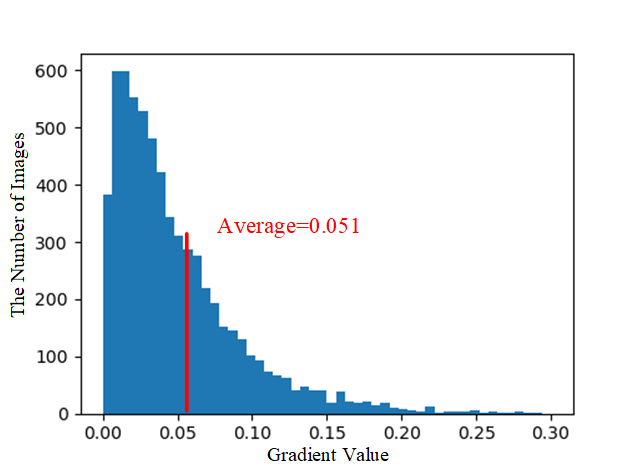}
	\caption{
		The statistical histogram of 6000 normal illumination images in the datasets \cite{Chen2018Retinex, wang2018gladnet, Lv2019AgLLNet}, where the average gradient value among all the selected images is $0.051$.
	}
	\label{avg}
\end{figure}

In this work, we design a gradient loss $\mathcal L_{gra}$ to desire the average gradient value of SLLEN's result to be an expected level $G$.
To find out such $G$, we calculate the average gradient values of 6000 normal illumination images in datasets \cite{Chen2018Retinex, wang2018gladnet, Lv2019AgLLNet}. The corresponding statistical histogram is illustrated in Figure~\ref{avg}.
It can be easily found from this figure that the average gradient value among all the selected images is $0.051$, which enables us initialize $G=0.051$ in this work.

\begin{figure*}[htb]
	\centering
	\includegraphics[scale=0.158]{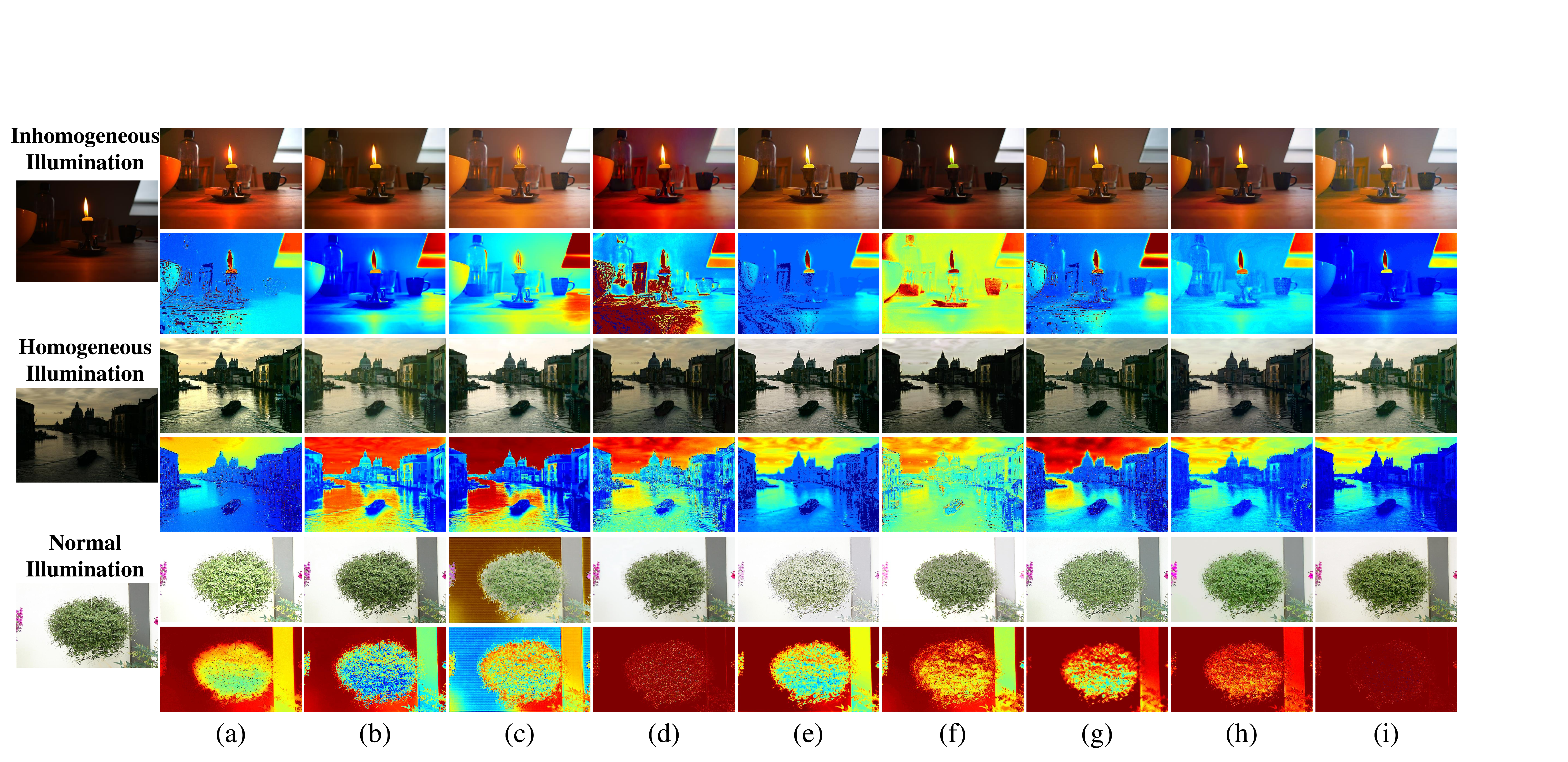}
	\caption{The comparison on the decomposed illumination maps between the proposed SLLEN and state-of-the-art LLE techniques. (a): LIME. (b): LD. (c): RF. (d): DCE++. (e): UNIE. (f): SCI. (g): ISSR. (h): SCL. (i): SLLEN.}
	\label{supplement_illumap}
\end{figure*}

% 23.2 / 0.80

\subsection{Comparing Decomposed Illumination Maps}
\label{section2}
During training, we design a loss function $\mathcal L_{itv}$ to ensure that illumination map $\bf{U}$ is as smooth as possible.
%	To prove the effectiveness of $\mathcal L_{itv}$, based on Retinex model, we calculate the decomposed $\bf{U}$ corresponding to the image enhanced by SLLEN and the state-of-the-art approaches, and compare their low-light image enhancement (LLE) performance. 
% To prove the effectiveness of $\mathcal L_{itv}$, based on Retinex model, we calculate the decomposed $\bf{U}$ using the enhanced results $\bf{O}$ of different approaches (i.e., $\bf{U}=\bf{I}/\bf{O}$, where $\bf{I}$ is low-light image). Meanwhile, we also compare their low-light image enhancement (LLE) performance. 
To prove the effectiveness of $\mathcal L_{itv}$, we calculate the decomposed $\bf{U}$ using the enhanced results of different approaches via Equation~\ref{eq:ReRetinex}. Meanwhile, we also compare their low-light image enhancement (LLE) quality.
The experimental results on three different kinds of examples (i.e., inhomogeneous illumination, homogeneous illumination, and normal illumination) are shown in Figure~\ref{supplement_illumap}.
%	To prove the effectiveness of $\mathcal L_{itv}$, based on Retinex model, we calculate the decomposed $\bf{U}$ using the enhanced results $\bf{O}$ of different approaches (i.e., $U=I/O$, where I is low-light image). Meanwhile, we also compare compare their low-light image enhancement (LLE) performance.
As seen from this figure, the proposed SLLEN can not only keep an excellent smoothness property, but also avoid over-exposure for the images with normal illumination.

\subsection{Ablation Study}% \label{Ablation}
\label{section3-1}
To check the efficacy of each component of SLLEN, we perform several ablation studies w.r.t. three loss functions, two feature enhancement blocks, and alternating training mechanism qualitatively and quantitatively. 
Here we use four commonly-used image quality metrics (i.e., peak signal-to-noise ratio (PSNR), structural similarity (SSIM) \cite{1284395}, lightness order error (LOE), and contrast enhancement based contrast-changed image quality (CEIQ) \cite{yan2019no}), to execute several quantitative ablation studies on the dataset LOL-test \cite{Chen2018Retinex}. 
\par
\subsubsection{Contribution of Each Loss Function} 
\begin{figure}[htbp]
	\centering
	\includegraphics[scale=0.155]{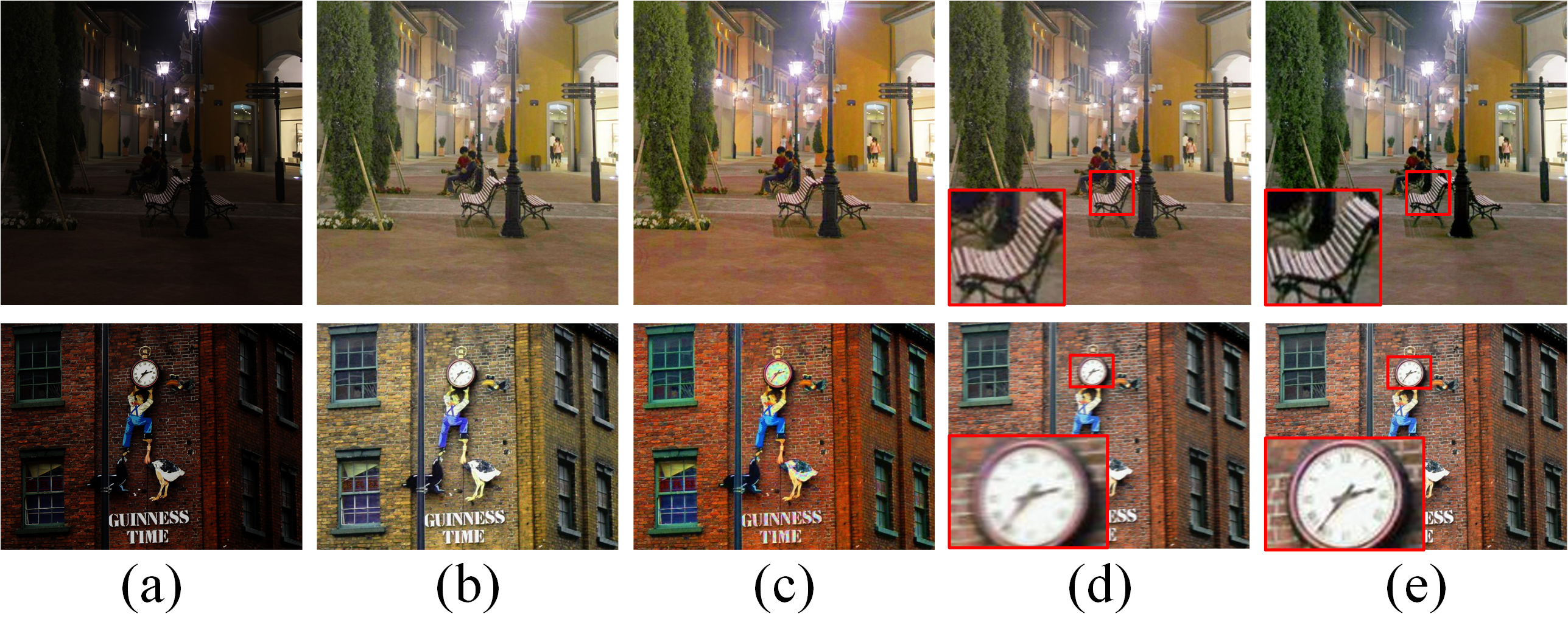}
	\caption{Ablation study of the contribution of each loss. (a): Inputs. (b) w/o $\mathcal L_{kd}$. (c) w/o $\mathcal L_{itv}$. (d) w/o $\mathcal L_{gra}$. (e) SLLEN.}
	%	 (Knowledge Distillation Loss $\mathcal L_{kd}$, Illumination Total Variation Loss $\mathcal L_{itv}$, and Gradient Loss $\mathcal L_{gra}$).
	\label{ablation_for_loss}
\end{figure}

We conduct SLLEN with various loss functions and present the corresponding results in Figure~\ref{ablation_for_loss}.
As seen, the enhanced results by SLLEN without $\mathcal L_{kd}$ are prone to slight color cast. The absence of $\mathcal L_{itv}$ tends to make the outputs suffer from visual unnaturalness. Moreover, removing $\mathcal L_{gra}$ makes SLLEN fail to restore the sharp contours. In contrast, the results enhanced by SLLEN with all the loss functions are superior regarding color authenticity, structure preservation, and edge sharpness. The comparing scores of SLLEN with various loss functions are illustrated in Table~\ref{Tab1}. As shown, SLLEN with all loss functions gives the best score, which means that the absence of each loss function leads to the deterioration of SLLEN's performance. %It is also verified that the loss of the $\mathcal L_{kd}$ would cause the decline of PSNR score, demonstrating $\mathcal L_{kd}$ is crucial to reviving the naturalness of enhanced images. The absence of the $\mathcal L_{itv}$ brings about the decline of SSIM score, meaning that $\mathcal L_{itv}$ plays an important role in structure preservation. The loss of the $\mathcal L_{gra}$ mainly induces the poor scores in terms of LOE and CEIQ, which demonstrates the ability of $\mathcal L_{gra}$ to restore the contrast of low-light images.
% Quantitative analysis of the contribution of each loss function can be found in the Appendix-\ref{section3}.

\setlength{\tabcolsep}{2mm}{}
\begin{table}\footnotesize
	\centering
	\caption{Quantitative ablation study of the contribution of each loss function, i.e., knowledge distillation loss $\mathcal L_{kd}$, illumination total variation loss $\mathcal L_{itv}$, and gradient loss $\mathcal L_{gra}$. Data in bold means the best and data in red indicates the second-best.}\
	\begin{tabular} {*{8}{c}}
		\toprule
		& Metric & w/o $\mathcal L_{kd}$ & w/o $\mathcal L_{itv}$ & w/o $\mathcal L_{gra}$ & SLLEN\\
		\midrule
		\multirow{3}{*}
		&\ PSNR$\uparrow$ & 22.4  & \JS{23.1}  & 22.7 & \textbf{23.8} \\
		&\ SSIM$\uparrow$ & 0.75 & 0.73 & \JS{0.80} & \textbf{0.84} \\
		&\ LOE$\downarrow$  & \JS{269} & 278 & 294 & \textbf{245} \\
		&\ CEIQ$\uparrow$ & 2.99 & \JS{3.08} & 2.91 & \textbf{3.22} \\
		\bottomrule
	\end{tabular}
	\label{Tab1}
\end{table}

\begin{figure}[htbp]
	\centering
	\includegraphics[scale=0.143]{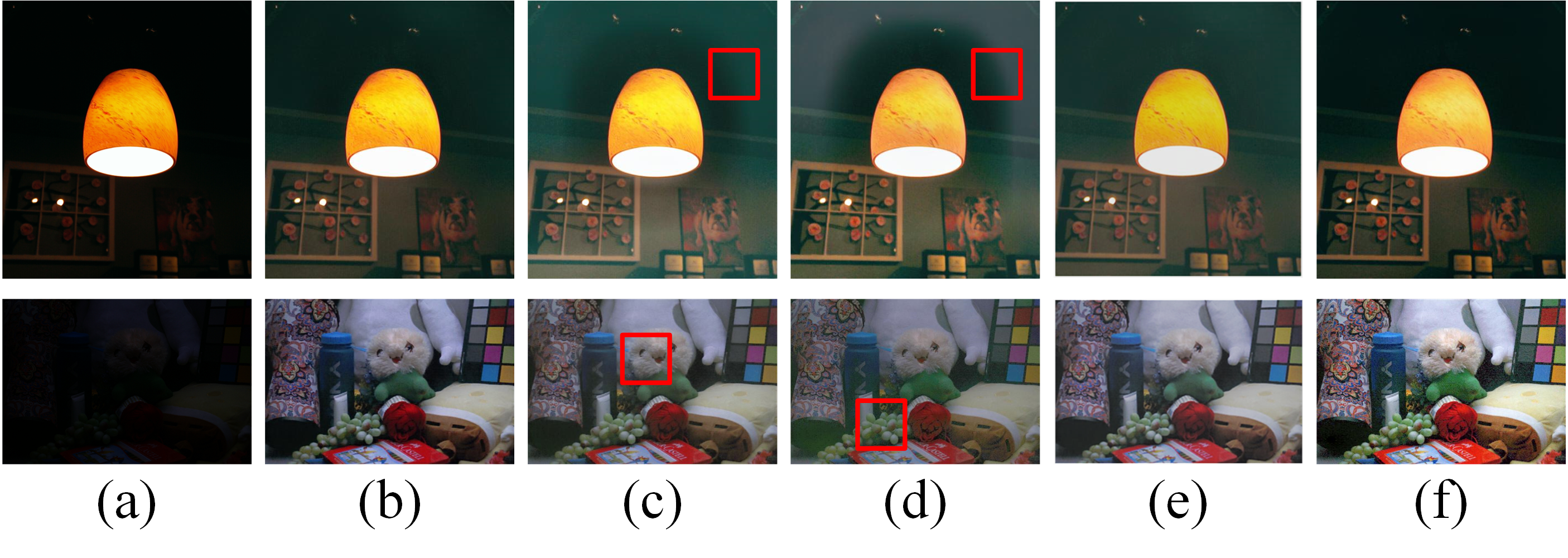}
	\caption{Ablation study of the contribution of each branch and alternating training mechanism. (a): Inputs. (b) SLLEN-1 (w/o HSBAB). (c) SLLEN-2 (w/o RSAEB). (d) SLLEN-3 (U-Net). (e) SLLEN-4 (w/o alternating training mechanism). (f) SLLEN.}
	\label{ablation_for_model}
\end{figure}

\subsubsection{Contribution of Each Feature Enhancement Block} Two feature enhancement blocks are included in the SLLEN, where the high-level semantic based attention block (HSBAB) is to use high-level semantic feature (HSF) $\bf{H}$ to guide the enhancement of low-level feature $\bf{L}$ and the random semantic aware enhancement block (RSAEB) is to impose intermediate embedding feature (IEF) $\bf{B}$ on low-level feature $\bf{L}$. %To check the efficacy of the two blocks in SLLEN, we implement an ablation study here. 
Here, we remove the HSBAB (introducing $\bf{H}$) and replace $\bf{L_{H}}$ with $\bf{L}$ as the input into the fusion block. For clarity, this ablated SLLEN is marked as SLLEN-1 (w/o HSBAB). Similarly, the RSAEB (introducing $\bf{B}$) is removed and $\bf{L}$ is fed into the fusion block in place of $\bf{L_{B}}$, and the corresponding ablated SLLEN is named as SLLEN-2 (w/o RSAEB). Finally, two blocks are both eliminated, which makes SLLEN degenerate into a traditional U-Net structure (SLLEN-3). An example illustration and the corresponding results enhanced by SLLEN-1, SLLEN-2, SLLEN-3, and SLLEN are shown in Figures~\ref{ablation_for_model}(a)$\sim$(d) and (f).

It can be seen from the figure that SLLEN-3 produces the worst results with white halos and low-contrast for the two given examples. Although SLLEN-2 is able to alleviate such problems, its results are still prone to poor visibility. In comparison, SLLEN and SLLEN-1 win the best and second best in terms of visual quality, respectively. This corroborates that the random IEF plays a more important role than HSF in improving network's performance. Table~\ref{Tab2} lists the scores of SLLEN-1, SLLEN-2, SLLEN-3, and SLLEN. It can be checked from the table that the performance of SLLEN is superior to others in terms of all the metrics, demonstrating that both enhancement blocks are indispensable. %It is worth mentioning that the absence of $\bf{H}$ only affects the LLE performance of SLLEN slightly, while removing $\bf{B}$ is bound to seriously deteriorate the LLE ability of SLLEN. This indicates $\bf{B}$ is dominant to the LLE performance of SLLEN compared to $\bf{H}$.

\setlength{\tabcolsep}{2mm}{}
\begin{table}\footnotesize
	\centering
	\caption{Quantitative ablation study of the contribution of each enhancement branch. Data in bold means the best and data in red indicates the second-best.}\
	\begin{tabular} {*{7}{c}}
		\toprule
		& Metric & SLLEN-1 & SLLEN-2 & SLLEN-3 & SLLEN-4 & SLLEN \\
		\midrule
		\multirow{3}{*}
		&\ PSNR$\uparrow$ & \JS{22.6} & 20.7 & 18.7 & 21.8 & \textbf{23.8} \\
		&\ SSIM$\uparrow$ & 0.76 & 0.68 & 0.60& \JS{0.78} & \textbf{0.84} \\
		&\ LOE$\downarrow$ & 281 & 357 & 489 & \JS{267} & \textbf{245} \\
		&\ CEIQ$\uparrow$ & \JS{3.19} & 2.89 & 2.78& 3.15 & \textbf{3.22} \\
		\bottomrule
	\end{tabular}
	\label{Tab2}
\end{table}
%generates compromised results with slight degradation aesthetically compared to the best output by SLLEN. This corroborates that the random IEF plays a more important role than HSF in network's performance. Quantitative analysis of the contribution of each branch can be found in the Supplementary Materials.

\subsubsection{Contribution of Alternating Training Mechanism}
%In this work, we proposed semantic-based auxiliary training mechanism, i.e., we design two networks (LLEmN and SSAS) sharing an encoder and train two networks by tuning the learnt parameters of shared encoder according to both SS training samples and LLE training samples. 
To investigate the affect of alternating training mechanism, we independently train LLEmN and SSaN with corresponding samples, and then combine them to implement semantic-aware LLE, such trained model is named as SLLEN-4. Figures~\ref{ablation_for_model}(e) and (f) show the experimental results of SLLEN-4 and SLLEN. It is obvious that the LLE performance of SLLEN is superior to that of SLLEN-4, which demonstrates the effectiveness of alternating training mechanism. The scores of SLLEN-4 and SLLEN listed in Table \ref{Tab2} also proves this conclusion.

\subsection{Benchmark Evaluations}
\label{section3-2}
\begin{figure*}[ht]
	\centering
	\includegraphics[scale=0.332]{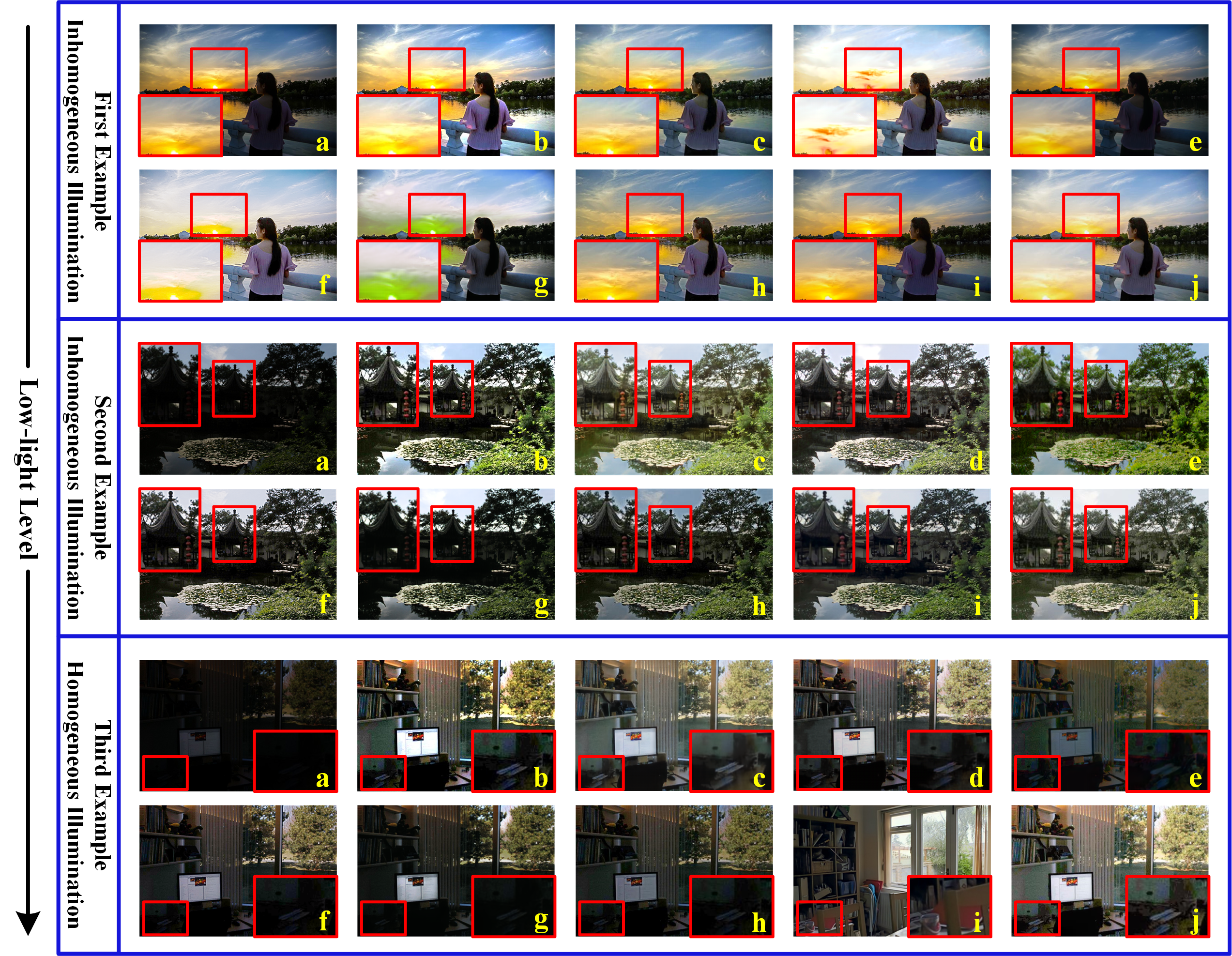}
	\caption{Qualitative comparison between the proposed SLLEN and state-of-the-art LLE techniques on three real-world examples. (a): Real-world images. (b): LIME. (c): LD. (d): RF. (e): DCE++. (f): UNIE. (g): SCI . (h): ISSR. (i): SCL. (j): SLLEN.}
	\label{without_label}
\end{figure*}
\begin{figure*}[ht]
	\centering
	\includegraphics[scale=0.2450]{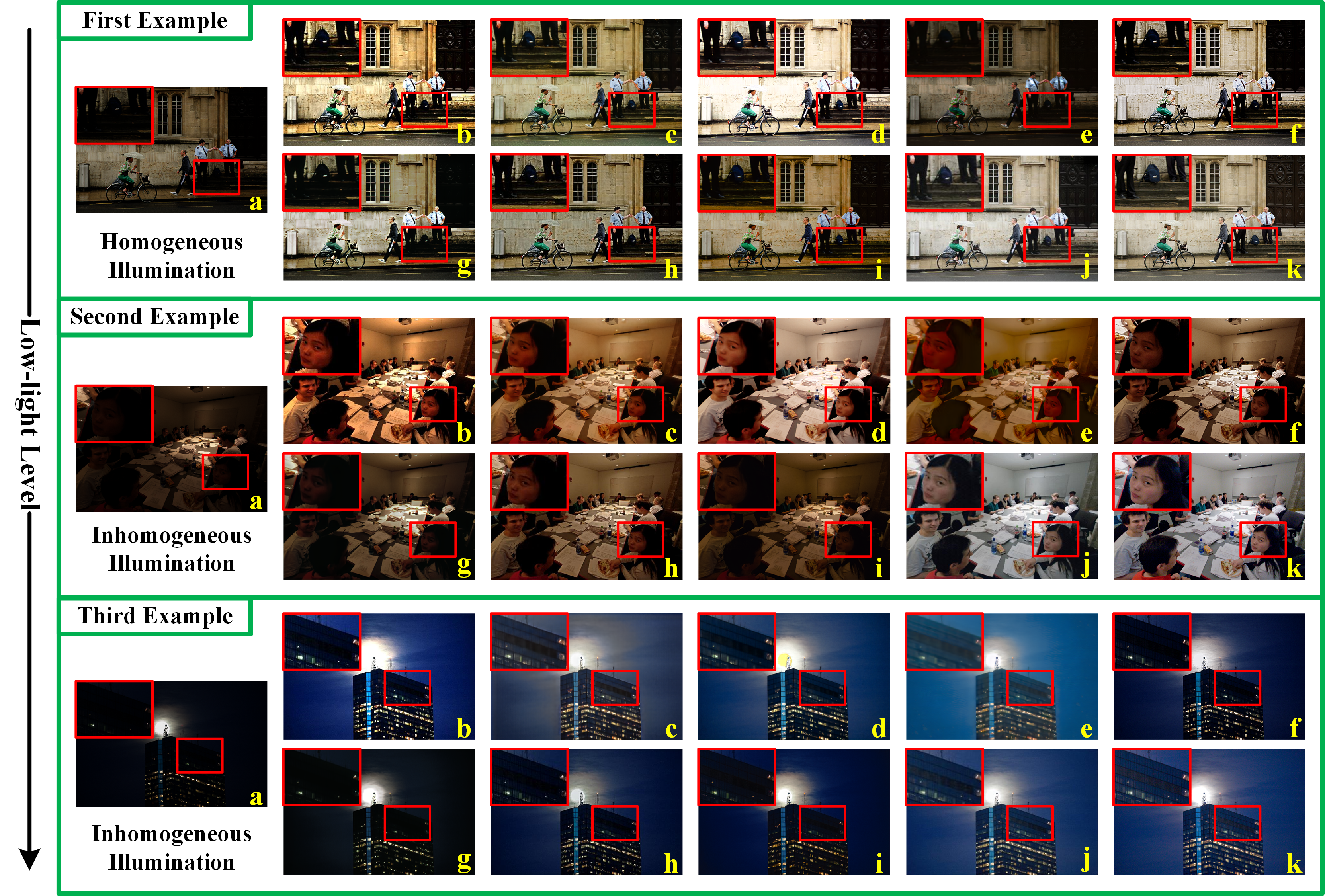}
	\caption{Qualitative comparison between the proposed SLLEN and state-of-the-art LLE techniques on three synthesis image examples.  (a): Synthesis images. (b): LIME. (c): LD. (d): RF. (e): DCE++. (f): UNIE. (g): SCI . (h): ISSR. (i): SCL. (j): SLLEN. (k): GT.}
	\label{with_label}
\end{figure*}
A series of experiments are conducted to investigate the performance of the proposed SLLEN and other state-of-the-art methods, including LIME (model-based, TIP 2016) \cite{guo2016lime}, LD (learning-based, ECCV 2024) \cite{Jiang_2024_ECCV}, RF (learning-based, ICCV 2023) \cite{retinexformer}, DCE++ (learning-based, TPAMI 2022) \cite{9369102}, UNIE (learning-based, ECCV 2022) \cite{jin2022unsupervised}, SCI (learning-based, CVPR 2022) \cite{Ma_2022_CVPR}, ISSR (semantic-guided learning-based, ACM-MM 2020) \cite{fan2020integrating}, SCL (semantic-guided learning-based, AAAI 2022) \cite{liang2022semantically}. More specifically, we perform both qualitative and quantitative experiments on challenging images picked from several benchmark datasets (GladNet-Dataset (paired) \cite{wang2018gladnet}, Synthetic Dataset (paired) \cite{Lv2019AgLLNet}, LOL-test (paired) \cite{Chen2018Retinex}, DICM (unpaired) \cite{lee2013contrast}, NPE (unpaired) \cite{wang2013naturalness}, and MEF (unpaired) \cite{ma2015perceptual}). %Besides, we further investigate the enhancement capability of proposed SLLEN and state-of-the-art methods on a high-level visual task.

\setlength{\tabcolsep}{1.3 mm}{}
\begin{table*}\footnotesize
	\centering
	\caption{Quantitative comparison between SLLEN and state-of-the-art methods. (Data in bold means the best and data in red indicates the second-best.)}\
	\begin{tabular} {*{20}{c}}
		\toprule
		Metrics & Database & LIME \cite{guo2016lime} & LD \cite{Jiang_2024_ECCV} & RF \cite{Zhang_2021_CVPR}  & DCE++ \cite{9369102} & UNIE  \cite{jin2022unsupervised} & SCI \cite{Ma_2022_CVPR}  & ISSR \cite{fan2020integrating}  & SCL \cite{liang2022semantically}  & SLLEN \\
		\midrule
		\multirow{3}{*}{PSNR$\uparrow$ / SSIM$\uparrow$}
		&\ GladNet   & 16.9 / 0.68  & \textbf{19.4} / 0.76 & 17.6 / \textcolor{red}{0.77} & 15.7 / 0.48 & 14.5 / 0.55 & 17.2 / 0.68  & 15.9 / 0.63 & 16.2 / 0.68 & \textcolor{red}{19.3} / \textbf{0.79} \\
		&\ Synthetic & 15.1 / 0.61  & 17.5 / 0.71 & \textcolor{red}{18.1} / \textcolor{red}{0.73} & 13.1 / 0.37 & 12.1 / 0.48 & 15.8 / 0.62 & 15.0 / 0.61 & 12.3 / 0.53 & \textbf{18.9} / \textbf{0.83} \\
		&\ LOL-test  & 17.1 / 0.67  & \textcolor{red}{21.7} / \textbf{0.84} & 17.7 / 0.74 & 15.0 / 0.45 & 21.4 / 0.75  & 13.8 / 0.53 & 12.4 / 0.47 & 13.3 / 0.61 & \textbf{23.8} / \textcolor{red}{0.83} \\
		\midrule
		\multicolumn{2}{c}{Average PSNR$\uparrow$ / Average SSIM$\uparrow$} & 16.4 / 0.65 & \textcolor{red}{19.5} / \textcolor{red}{0.77} & 17.8 / 0.75 & 14.6 / 0.43 & 16.0 / 0.59 & 15.6 / 0.61 & 14.4 / 0.57 & 13.9 / 0.61 & \textbf{20.7} / \textbf{0.82} \\
		
		\midrule
		\multirow{3}{*}{LOE$\downarrow$ / CEIQ$\uparrow$}
		&\  DICM & 832 / \JS{3.30} &	\textbf{236} / 3.29 &	439 / 3.18 & 	502 / 3.13 & 	\JS{277} / 3.26 &	362 / 3.01 &	521 / 3.00 &	392 / 3.10 &	284 / \textbf{3.40} \\
		&\  NPE  & 852 / \JS{3.44} &	\textbf{261} / 3.10 &   382 / 3.31 & 	571 / 3.37 & 	309 / 3.27 &	346 / 3.34 &	421 / 3.27 &	393 / 3.33 &	\JS{269} / \textbf{3.61} \\
		&\  MEF  & 603 / \JS{3.37} &	271 / 2.90 &	246 / 3.29 &  	459 / 3.31 & 	264 / 3.32 & 	\JS{202} / 3.18 & 	246 / 2.78 & 	296 / 3.14 & 	\textbf{195} / \textbf{3.53} \\
		\midrule
		\multicolumn{2}{c}{Average LOE$\downarrow$ / Average CEIQ$\uparrow$} & 762 / \JS{3.37} & \JS{256} / 3.10 & 356 / 3.26 & 511 / 3.27 & 283 / 3.28 & 303 / 3.18 & 396 / 3.02 & 360 / 3.19 & \textbf{249} / \textbf{3.51} \\

		\bottomrule
	\end{tabular}
	\label{quantitative}
\end{table*}
%  from Ref. \cite{7120119}

\subsubsection{Qualitative Comparisons}
%\textbf{Qualitative Comparison on Real-World Images.}
To assess the LLE effects of SLLEN, three challenging real-world images are selected to facilitate visual and perceptual comparisons of different techniques. Both challenging low-light images and the experimental results are illustrated in Figure~\ref{without_label}. It is observed from Figures~\ref{without_label}(b), and (g) that LIME, and SCI easily lead to over-restoration especially in the bright regions (see the second example).
Figure~\ref{without_label}(e) shows that DCE++ tends to induce low contrast in its results (see the second example).
As seen in Figures~\ref{without_label}(f), (h), and (i), although UNIE, ISSR, and SCL are able to light up the most regions, the rest part of the enhanced results still remains dim (see the third example).
By comparison, as seen in Figure~\ref{without_label}(j), SLLEN can work well on low-light images with inhomogeneous illumination and restore richer textures (Best viewed by zooming in). 

\begin{figure*}[htbp]
	\centering
	\includegraphics[scale=0.140]{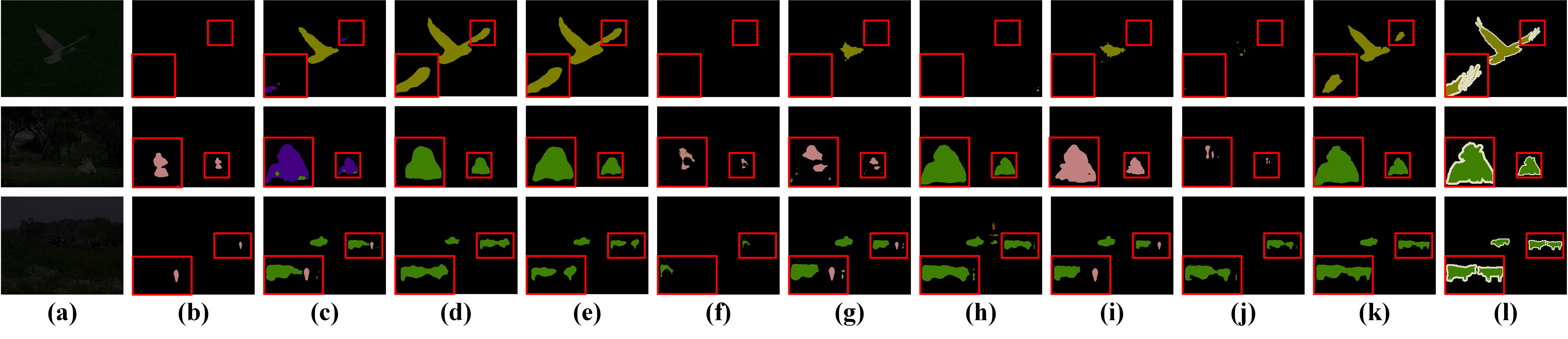}
	\caption{Two examples of semantic segmentation results on the low-light images and enhanced outputs by different algorithms.  (a): Low-light images. (b): Results of (a). (c): LIME. (d): LD. (e): RF. (f): DCE++. (g): UNIE. (h): SCI. (i): ISSR. (j): SCL. (k): SLLEN. (l): GT.}
	\label{semantic_res}
\end{figure*}

\normalsize

%\subsubsection{Qualitative and Quantitative Comparisons}
%\subsubsection{Qualitative Comparison on Synthesis Images.}
%\textbf{Qualitative Comparison on Synthesis Images.}
In addition, we also evaluate the restoration quality of the proposed SLLEN and the state-of-the-art approaches on three reference datasets, i.e., GladNet-Dataset \cite{wang2018gladnet}, Synthetic Dataset \cite{Lv2019AgLLNet}, and LOL-test \cite{Chen2018Retinex}. Three low-light examples are picked to facilitate the comparison. The corresponding results are illustrated in Figure~\ref{with_label}. In general, the results in Figure~\ref{with_label} are similar to those for real-world images illustrated in Figure~\ref{without_label}. Most of the competitors lack the ability to enhance the dim part for the example with non-uniform illumination. On the contrary, it can highlight that the results in Figure~\ref{with_label}(j) enhanced by SLLEN are more natural and satisfactory aesthetically with reference to the ground truth images (GT).

\subsubsection{Quantitative Comparison}
To overcome the bias caused by subjective assessment, we implement quantitative comparison using two benchmark reference metrics (PSNR and SSIM \cite{1284395}), and two non-reference metrics (LOE and CEIQ \cite{yan2019no}).
Table~\ref{quantitative} gives the average scores of PSNR and SSIM on three paired datasets (GladNet-Dataset \cite{wang2018gladnet}, Synthetic Dataset \cite{Lv2019AgLLNet}, and LOL-test \cite{Chen2018Retinex}), and the average scores of LOE and CEIQ on three unpaired datasets (DICM \cite{lee2013contrast}, NPE \cite{wang2013naturalness}, and MEF \cite{ma2015perceptual}).
It can be found from this table that SLLEN achieves the best average scores in terms of SSIM and PSNR, which indicates that the results of SLLEN are more similar to the GT compared to those of other state-of-the-arts. Moreover, the advantages of SLLEN in LOE and CEIQ further demonstrate its superiority regarding illumination correction and contrast restoration.

\subsubsection{User Study}
We also perform a user study to assess the subjective LLE ability of various techniques. 
We process various low-light images from the datasets (DICM \cite{lee2013contrast}, NPE \cite{wang2013naturalness}, MEF \cite{ma2015perceptual}, LIME \cite{guo2016lime}, and VV\footnote{https://sites.google.com/site/vonikakis/datasets}) by different methods. 
For each low-light image, we display it on a screen and provide the outputs from the different competitors. 
We invite 10 human subjects to score the perceptual quality of the enhanced images. 
The subjects are required to consider: 1) whether the results are over-/under-exposed; 2) whether the results are pleasing aesthetically; 3) whether the results produce color deviation. 
The scores of perceptual quality range from $1$ to $5$ (worst to best quality). 
As summarized in Table~\ref{userstudy}, SLLEN wins the best average US score on the given datasets.

\begin{table*}\footnotesize
	\centering
	\caption{The score of the user study (US)$\uparrow$ on the datasets (DICM \cite{lee2013contrast}, NPE \cite{wang2013naturalness}, MEF \cite{ma2015perceptual}, LIME \cite{guo2016lime}, and VV) processed by different methods. Data in bold means the best and data in red indicates the second-best.}\
	\begin{tabular} {*{12}{l}}
		\toprule
		& Database & LIME \cite{guo2016lime} & RUAS \cite{liu2021retinex} & LTCLL \cite{Zhang_2021_CVPR}  & DCE++ \cite{9369102} & UNIE  \cite{jin2022unsupervised} & SCI \cite{Ma_2022_CVPR}  & ISSR \cite{fan2020integrating}  & SCL \cite{liang2022semantically}  & SLLEN \\
		
		\midrule
		
		& DICM \cite{lee2013contrast} & \JS{3.868} & 3.012 & 3.415 & 3.376 & 2.981 & 2.417 & 2.639 & 3.604 & \textbf{3.988} \\
		& NPE \cite{wang2013naturalness} & 3.312 & 3.055 & 3.038 & \JS{3.861} & 3.524 & 3.226 & 2.811 & 3.426 & \textbf{4.014} \\
		& MEF \cite{ma2015perceptual} & 3.451 & 3.359 & 3.004 & \JS{3.953} & 3.212 & 3.147 & 2.908 & 3.722 & \textbf{3.971} \\
		& LIME \cite{guo2016lime} & 3.056 & 2.983 & 3.150 & \JS{3.418} & 2.447 & 2.732 & 2.657 & 3.331 & \textbf{3.726} \\
		& VV & 3.401 & 3.002 & 3.185 & \JS{3.767} & 2.819 & 2.435 & 2.767 & 3.456 & \textbf{3.871} \\
		\midrule
		& Average & 3.418 & 3.082 & 3.158 & \JS{3.675} & 2.997 & 2.791 & 2.756 & 3.508 & \textbf{3.914} \\
		\bottomrule
	\end{tabular}
	\label{userstudy}
\end{table*}

\subsection{Semantic Segmentation in the Dark}
\label{section3-3}
We further evaluate the performance of state-of-the-art LLE techniques on semantic segmentation task under low-light conditions.
In specific, we randomly select 100 images from visual object classes (VOC) \cite{Everingham10} and darken them according to \cite{Zhang_2021_CVPR}. These darkened images are then enhanced by the aforementioned comparable methods and our SLLEN. Finally, we adopt \cite{chen2018encoder} as a baseline to rank the segmentation capability for different algorithms.
Figure~\ref{semantic_res} gives the semantic results on three given examples.
As shown, the semantic maps of output results of SLLEN are the closest to the GT compared to those of its competitors.

\setlength{\tabcolsep}{1.6mm}{}
\begin{table*}\footnotesize
	\centering
	\caption{The average mIoU$\uparrow$ scores of semantic segmentation results on the specific detected objects over 100 corresponding outputs enhanced by different algorithms on the dataset Visual Object Classes (VOC) \cite{Everingham10}. Data in bold means the best and data in red indicates the second-best.}\
	\begin{tabular} {*{12}{l}}
		\toprule
		&Detected Objects & LIME \cite{guo2016lime} & RUAS \cite{liu2021retinex} & LTCLL \cite{Zhang_2021_CVPR}  & DCE++ \cite{9369102} & UNIE  \cite{jin2022unsupervised} & SCI \cite{Ma_2022_CVPR}  & ISSR \cite{fan2020integrating}  & SCL \cite{liang2022semantically}  & SLLEN \\
		\midrule
		&Background    &  85.93 &	86.38 &	86.33 &	86.19 &	85.21 &	86.61 &	87.16 &	87.04 & 88.08  \\
		&Aeroplane     &  71.52 &	59.26 &	49.41 &	55.08 &	49.67 &	65.12 &	55.03 &	72.19 &	70.27 \\
		&Bicycle       &  52.07 &	37.38 &	46.93 &	45.88 &	23.76 &	51.54 &	48.70 &	47.73 &	51.89 \\
		&Bird          &  92.64 &	92.94 &	93.94 &	81.47 &	92.22 &	93.22 &	92.13 &	92.96 &	92.88 \\
		&Boat          &  28.50 &	34.77 &	11.10 &	24.55 &	39.71 &	39.16 &	46.15 &	43.42 &	41.75 \\
		&Bottle        &  49.79 &	45.52 &	43.20 &	44.22 &	51.01 &	47.60 &	51.42 &	47.33 &	62.17 \\
		&Bus           &  79.85 &	78.74 &	80.85 &	82.13 &	73.97 &	81.57 &	77.94 &	80.67 &	86.56 \\
		&Car           &  81.29 &	85.51 &	62.63 &	80.14 &	67.45 &	85.93 &	82.61 &	84.65 &	87.13 \\
		&Cat           &  84.20 &	83.83 &	56.80 &	78.23 &	85.55 &	85.38 &	82.74 &	84.91 &	86.79 \\
		&Chair         &  8.28  &   7.30 &	5.83  &  6.55 & 6.70  &	7.84  &	10.12 &	8.61 &	16.59 \\
		&Cow           &  73.91 &	79.52 &	78.08 &	74.37 &	76.37 &	79.75 &	81.52 &	77.92 &	89.43 \\
		&Diningtable   &  55.98 &	56.89 &	74.35 &	71.76 &	53.80 &	59.57 &	64.21 &	59.90 &	73.38 \\
		&Dog           &  49.77 &	49.78 &	34.29 &	54.82 &	37.91 &	58.99 &	53.17 &	59.44 &	70.76 \\
		&Horse         &  53.50 &	58.27 &	59.35 &	57.92 &	55.11 &	56.24 &	54.90 &	57.94 &	65.19 \\
		&Motorbike     &  76.90 &	71.50 &	76.42 &	79.12 &	56.27 &	81.02 &	79.40 &	73.45 &	81.89 \\
		&Person        &  69.80 &	67.85 &	61.03 &	69.49 &	66.93 &	70.61 &	69.93 &	70.57 &	75.45 \\
		&Pottedplant   &  23.76 &	11.86 &	10.97 &	19.62 &	7.31  &	17.77 &	18.75 &	22.16 &	29.84 \\
		&Sheep         &  63.11 &	86.80 &	87.12 &	78.01 &	85.58 &	87.17 &	85.65 &	87.36 &	87.94 \\
		&Sofa          &  20.43 &	30.87 &	14.42 &	21.34 &	23.90 &	24.73 &	35.68 &	16.46 &	28.58 \\
		&Tvmonitor     &  11.09 &	17.98 &	3.28  & 10.10 & 5.31  &	11.98 &	21.58 &	8.42 &	23.17 \\
		\midrule
		&Average       & 56.62 & 57.15 & 51.82 & 56.05 & 52.19 & 59.59 & \JS{59.94} & 59.16 & \textbf{65.49} \\
		\bottomrule
	\end{tabular}
	\label{Tab3}
\end{table*}

\setlength{\tabcolsep}{4mm}{}
\begin{table}\footnotesize
	\centering
	\caption{Running time comparison (in second) of the proposed SLLEN and different state-of-the-art methods.}\
	\begin{tabular} {*{12}{l}}
		\toprule
		& Method                           & Running time & \quad Platform \\
		
		\midrule
		
		& LIME \cite{guo2016lime}          &  0.05002       & MATLAB (CPU) \\                                                
		& LD \cite{Jiang_2024_ECCV}       &  0.29631   & Pytorch (GPU) \\                                                        
		& RF \cite{retinexformer}     &  0.36778   & Pytorch (GPU)   \\                                             
		& DCE++ \cite{9369102}             &  0.00057   & Pytorch (GPU)   \\                                                
		& UNIE \cite{jin2022unsupervised}  &  0.00462   & Pytorch (GPU)   \\                                               
		& SCI \cite{Ma_2022_CVPR}          &  0.00034   & Pytorch (GPU)   \\                                       
		& ISSR \cite{fan2020integrating}   & 5.90612  & TensorFlow (GPU)   \\                                            
		& SCL \cite{liang2022semantically} &  0.00047   & Pytorch (GPU)   \\                                                
		& SLLEN                            &  0.00403  & Pytorch (GPU)  \\
		\bottomrule
	\end{tabular}
	\label{runtime}
\end{table}

%\begin{table*}\footnotesize
%	\centering
%	\caption{The average mIoU scores of semantic segmentation results on the 100 corresponding outputs enhanced by different algorithms.}\
%	\begin{tabular} {*{12}{c}}
%		\toprule
%		&Metrics & Database & LIME \cite{guo2016lime} & RUAS \cite{liu2021retinex} & LTCLL \cite{Zhang_2021_CVPR}  & DCE++ \cite{9369102} & UNIE  \cite{jin2022unsupervised} & SCI %\cite{Ma_2022_CVPR}  & ISSR \cite{fan2020integrating}  & SCL \cite{liang2022semantically}  & SLLEN \\
%		\midrule
%		&\ mIoU\%$\uparrow$ & 100 darkened images in VOC &  56.62 & 57.15 & 51.82 & 56.05 & 52.19 & 59.59  & 59.94  & 59.16 & 61.15 \\
%		\bottomrule
%	\end{tabular}
%	\label{semantic}
%\end{table*}
To further facilitate the comparison, mean Intersection over Union (mIoU), whose larger value means better semantic segmentation result, is used to evaluate the performance of different algorithms on semantic segmentation. The corresponding average mIoU scores on the 100 examples by each algorithm are reported in Table~\ref{Tab3}. As seen, the score of SLLEN outperforms those of the eight comparable approaches.

\subsection{Limitation and Broader Impacts}
\label{section6}

Table~\ref{runtime} reports the running time of the different techniques averaged on 32 images of size $512\times512\times3$, which is measured on a PC with an Nvidia RTX 3080Ti GPU and Intel I9 12900KF CPU. 
Due to the integration of semantic segmentation auxiliary-network, the proposed SLLEN does not show an advantage in computational efficiency compared to some algorithms.
However, the LLE quality and semantic extraction performance of SLLEN are undoubtedly the best (see Figures~\ref{supplement_illumap},~\ref{without_label},~\ref{with_label},~\ref{semantic_res} and Tables~\ref{userstudy}, \ref{Tab3}). Therefore, our SLLEN can serve as an excellent candidate that provides both high-quality LLE and semantic extraction services. 
%Moreover, to the best of our knowledge, we are the first to introduce the random intermediate embedding feature, i.e., the information extracted from the intermediate layer of semantic segmentation network into a unified framework for LLE. This can enlighten researchers to explore more solutions for how to leverage the latent relationships between the low-level feature and semantic feature to achieve a promising LLE. 

To the best of our knowledge, we are the first to design a training strategy with shared encoder to make an interaction between LLE and SS tasks. Moreover, the intermediate embedding feature containing random semantic maps that are different from each other, i.e., the information extracted from the intermediate layer of semantic segmentation network (SSaN), is also introduced into a unified framework for better LLE. Such idea can enlighten researchers to explore more solutions for how to better leverage the latent relationships among different tasks to achieve a higher performance. In the future, we will try to adopt several model-optimization techniques, such as knowledge transfer, knowledge distillation, and model pruning, to speed up or simplify the proposed network.

\section{Conclusion}
In this work, we develop a semantic-aware LLE network (SLLEN), which consists of LLE main-network (LLEmN) and SS auxiliary-network (SSaN). 
The SSaN, sharing an encoder with LLEmN, is proposed to accomplish the semantic segmentation task, which can provide semantic information (HSF and IEF) for LLEmN.
The LLEmN is a two-branch LLE network; for the first branch, an attention mechanism is utilized to integrate HSF into low-level feature; for the second branch, we estimate the two vectors from IEF to drive the adjustment of low-level feature. 
%
% Among the provided semantic features, HSF is directly extracted from the output map of AS and IEF is grasped from the intermediate layer of SSN. 
The two adjusted features from two branches are fused together to be decoded for the final enhanced result. 
%
% Unlike the other comparable works, our SLLEN makes efficient use of random semantic information in IEF, making low-level feature and semantic feature well interacted. 
%
Furthermore, to fully facilitate the interaction of SS and LLE tasks, we design an alternating training mechanism to iteratively train LLEmN and SSaN, by repeatedly tuning the learnt parameters in the shared encoder.
% which facilitate the interaction between the low-level feature and high-level (semantic) feature.
%
Extensive experiments demonstrate that the proposed SLLEN performs better on various synthetic datasets and real-world scenes.

\IEEEpeerreviewmaketitle

\ifCLASSOPTIONcaptionsoff
\newpage
\fi

\bibliographystyle{IEEEtran}
\bibliography{ref}

\end{document}